\documentclass[journal]{IEEEtran}

\ifCLASSINFOpdf
  \usepackage[pdftex]{graphicx}
  \graphicspath{{figures/}{pictures/}{images/}{./}}
  \DeclareGraphicsExtensions{.pdf,.jpeg,.png,.jpg}
\else
  \usepackage[dvips]{graphicx}
  \graphicspath{{../eps/}}
  \DeclareGraphicsExtensions{.eps}
\fi

\usepackage{amsmath}
\usepackage{amssymb} 

\usepackage{url}
\usepackage{booktabs} 
\usepackage{enumitem}
\setlist{nosep}
\usepackage{xcolor}

\usepackage[hidelinks]{hyperref}
\usepackage{orcidlink}

\usepackage{fvextra}
\usepackage{mdframed}
\mdfdefinestyle{promptbox}{
  linecolor=black!35, linewidth=0.6pt,
  backgroundcolor=black!4,
  innerleftmargin=8pt, innerrightmargin=8pt,
  innertopmargin=6pt,  innerbottommargin=6pt,
  skipabove=6pt, skipbelow=6pt,
}

\newcommand{\mycomment}[1]{{ #1}}
\newcommand{\tablecomment}{}

\begin{document}
%
\title{Shape2Animal: Creative Animal Generation from Natural Silhouettes}

\author{
    Quoc-Duy Tran\orcidlink{0009-0003-1773-1520}\textsuperscript{1,2},
    Anh-Tuan Vo\orcidlink{0009-0002-3547-9907
    }\textsuperscript{1,2},
    Dinh-Khoi Vo\orcidlink{0000-0001-8831-8846}\textsuperscript{1,2},
    Tam V. Nguyen\orcidlink{0000-0003-0236-7992}\textsuperscript{3},
    Minh-Triet Tran\orcidlink{0000-0003-3046-3041}\textsuperscript{1,2}, 
    Trung-Nghia Le\orcidlink{0000-0002-7363-2610}\textsuperscript{1,2}*\thanks{*Corresponding author. Email: ltnghia@fit.hcmus.edu.vn} \\ 
    \textsuperscript{1}University of Science, Ho Chi Minh City, Vietnam \\
    \textsuperscript{2}Vietnam National University, Ho Chi Minh City, Vietnam \\
    \textsuperscript{3}University of Dayton, Ohio, US 
}


\markboth{IEEE MultiMedia Submission}%
{Tran \MakeLowercase{\textit{et al.}}: Shape2Animal: Creative Animal Generation from Natural Silhouettes}

\maketitle

\begin{abstract}
Humans possess a unique ability to perceive meaningful patterns in ambiguous stimuli, a cognitive phenomenon known as pareidolia. This paper introduces Shape2Animal framework to mimics this imaginative capacity by reinterpreting natural object silhouettes, such as clouds, stones, or flames, as plausible animal forms. Our automated framework first performs open-vocabulary segmentation to extract object silhouette and interprets semantically appropriate animal concepts using vision-language models. It then synthesizes an animal image that conforms to the input shape, leveraging text-to-image diffusion model and seamlessly blends it into the original scene to generate visually coherent and spatially consistent compositions. We evaluated Shape2Animal on a diverse set of real-world inputs, demonstrating its robustness and creative potential. Our Shape2Animal can offer new opportunities for visual storytelling, educational content, digital art, and interactive media design.
\end{abstract}

\begin{IEEEkeywords}
Pareidolia, Open-Vocabulary Segmentation, Silhouettes-Guided Interpretation, Text-to-Image Animal Synthesis.
\end{IEEEkeywords}


%
\IEEEpeerreviewmaketitle

\begin{figure*}[!t]
  \centering
  \includegraphics[width=\textwidth]{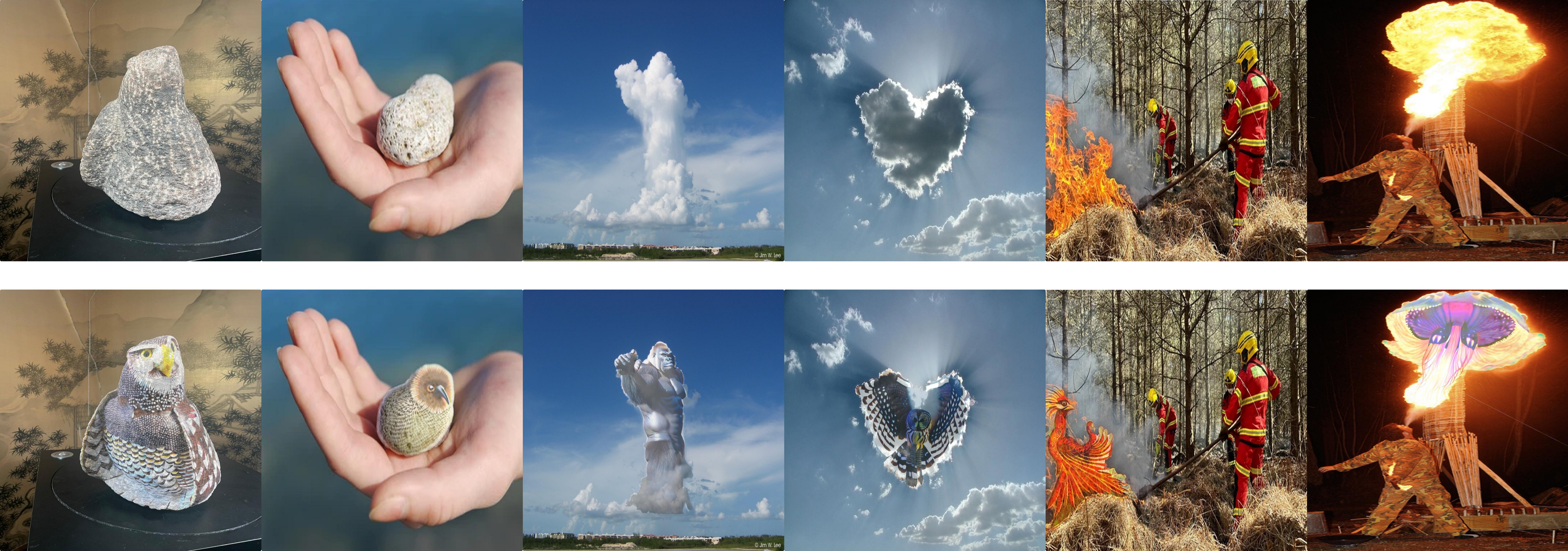}
  \caption{Shape2Animal emulates the human ability of pareidolia, reimagining natural silhouettes as animals by leveraging cutting-edge vision-language technologies to bring imagination to life within real-world scenes.}
  \label{fig:teaser}
\end{figure*}

\section{Introduction}
\IEEEPARstart{H}{uman} perception is remarkably adept at extracting meaning from ambiguity, a capacity exemplified by \textit{pareidolia}, the cognitive phenomenon in which individuals perceive familiar forms within vague or abstract stimuli~\cite{Bednarik2024CollectiveP, Liu2014FacePareidolia}. Whether seeing animals in clouds, faces in rocks, or creatures in flames, pareidolia reveals not only the brain's deep-seated pattern recognition mechanisms but also its imaginative flair~\cite{Doniger2012AnimalsImagination, MartinezConde2015BrainSeesFaces}. This interplay between perception and projection reflects a uniquely human tendency to impose semantic structure on randomness, a trait that has long inspired art, mythology, and visual storytelling~\cite{Ramachandran1999ArtBrain}.

In computational design, emulating pareidolia opens new creative frontiers where machines not only analyze the visual world but reimagine it~\cite{Somaini2022OnTA}. Inspired by this perceptual phenomenon, we investigate how AI systems can collaborate with human-like intuition to transform abstract shapes into meaningful interpretations, turning ambiguity into intentional design. Specifically, reconstructing natural object silhouettes as plausible animal forms, while maintaining geometric structure and contextual coherence, bridges the gap between human cognitive creativity and computational image synthesis. This paradigm enables new forms of visual content generation with applications in storytelling, digital art, education, and interactive media. Despite recent progress in generative modeling, current AI systems remain limited in their ability to perform this task.  
Text-to-image diffusion models~\cite{podell2023sdxlimprovinglatentdiffusion, saharia2022photorealistictexttoimagediffusionmodels} lack fine-grained spatial control over arbitrary masks and rely heavily on textual prompts, limiting their ability to infer meaning directly from shapes. Image-to-image translation techniques~\cite{zhu2020unpairedimagetoimagetranslationusing} often require paired training data or domain-specific customization, which reduces their adaptability. Even enhanced methods like ControlNet~\cite{zhang2023addingconditionalcontroltexttoimage} emphasize geometric precision but fall short of performing semantic reasoning on ambiguous forms. These limitations create a critical research gap in autonomous shape-driven generative systems that blend structural fidelity with creative reinterpretation.

To address the gap between human cognitive creativity and the limitations of current generative models, we propose Shape2Animal, a fully automated framework that leverages recent advances in vision-language technologies to synthesize animal imagery from natural object silhouettes (Fig.~\ref{fig:teaser}). Unlike prior approaches that rely on predefined prompts or rigid input formats, Shape2Animal interprets natural silhouettes as semantic cues, enabling the generation of animal forms that are both visually coherent and shape-consistent. It leverages Grounding DINO~\cite{Zhang:2023:GroundingDINO} and SAM~\cite{Kirillov:2023:SAM} for flexible, open-set segmentation, allowing it to extract ambiguous natural silhouettes from complex scenes. Semantic interpretation is performed using Gemini 2.5\footnote{\url{https://ai.google.dev/gemini-api/docs/models\#gemini-2.5-flash-preview}}, a multimodal large language model that generates meaningful animal concepts based solely on visual shape cues, resulting in dynamic prompts. To ensure that the synthesized imagery adheres to the original silhouette and respects scene geometry, the system incorporates Stable Diffusion XL~\cite{podell2023sdxlimprovinglatentdiffusion} guided by ControlNet depth maps. Synthesized animal images are seamlessly blended into the original scene to generate visually coherent and spatially consistent compositions.

To evaluate the effectiveness of the proposed framework, we conducted a comprehensive assessment and a user studies on a dataset of curated natural object images encompassing diverse categories such as clouds, stones, and fire. The results highlight Shape2Animal's ability to produce visually coherent and creatively engaging outputs, and its potential for shape-guided image manipulation. Our Shape2Animal can offer a versatile tool for artists, designers, and creative practitioners seeking to push the boundaries of automated visual content creation. In summary, our main contributions are as follows: 

\begin{itemize} 
\item We propose Shape2Animal, a novel end-to-end framework for silhouette-driven animal synthesis from natural scenes.

\item We introduce a simple yet efficient solution for interpreting ambiguous natural silhouettes as animal forms. Open-set segmentation is combined with multimodal large language model for semantic reasoning to generate meaningful animal concepts directly from visual shape cues.

\item We integrate Stable Diffusion XL with ControlNet depth maps, aligning the generated imagery with the original silhouette and scene geometry.

\item Comprehensive analysis across diverse input scenarios without manual intervention show the creative potential of our method. 
\end{itemize}

\section{Related Work}

Pareidolia, the human tendency to perceive familiar forms such as faces or animals in ambiguous stimuli, has been extensively studied in cognitive neuroscience and psychology~\cite{Liu2014FacePareidolia, MartinezConde2015BrainSeesFaces, Doniger2012AnimalsImagination}, yet remains largely underexplored in computational creativity. While some prior work has touched on artistic or perceptual simulations of pareidolia, few generative models have aimed to emulate this human-like reinterpretation of abstract shapes. 

In parallel, the field of image synthesis has advanced significantly, beginning with GANs~\cite{goodfellow2014generativeadversarialnetworks}, and evolving into text-to-image diffusion models such as Stable Diffusion~\cite{Rombach_2022_CVPR}, Imagen~\cite{saharia2022photorealistictexttoimagediffusionmodels}, and DALL·E 2~\cite{ramesh2022hierarchicaltexttoimage}, which generate diverse and high-quality images from textual descriptions. However, these models rely on explicit prompts and do not infer semantic meaning directly from visual stimuli, particularly ambiguous or naturally occurring shapes. Some methods ~\cite{zhang2023addingconditionalcontroltexttoimage, mou2023t2iadapter} were developed to enhance control synthesis through structural guidance, such as masks, edges, or depth maps. Nevertheless, they remain fundamentally prompt-driven and are not designed for semantic reinterpretation of unstructured visual input. 

On the other hand, vision-language models like CLIP~\cite{radford2021learningtransferable}, Flamingo~\cite{alayrac2022flamingofewshot}, and the more recent Gemini 2.5 have demonstrated strong visual reasoning capabilities by bridging visual and textual representations, but are rarely employed to drive generative synthesis from visual-only cues. 

In contrast to prior methods, our work introduces a novel pareidolia-inspired image generation task: transforming ambiguous natural silhouettes into plausible animal forms without textual input. By leveraging open-set segmentation, semantic reasoning from visual input, and spatially guided synthesis, our method bridges perception, imagination, and generation, moving beyond traditional prompt-based diffusion pipelines toward shape-driven, cognitively inspired visual content creation.

\section{Proposed Method}

\subsection{Overview}

\begin{figure*}[t!] 
    \centering
\includegraphics[width=\textwidth]{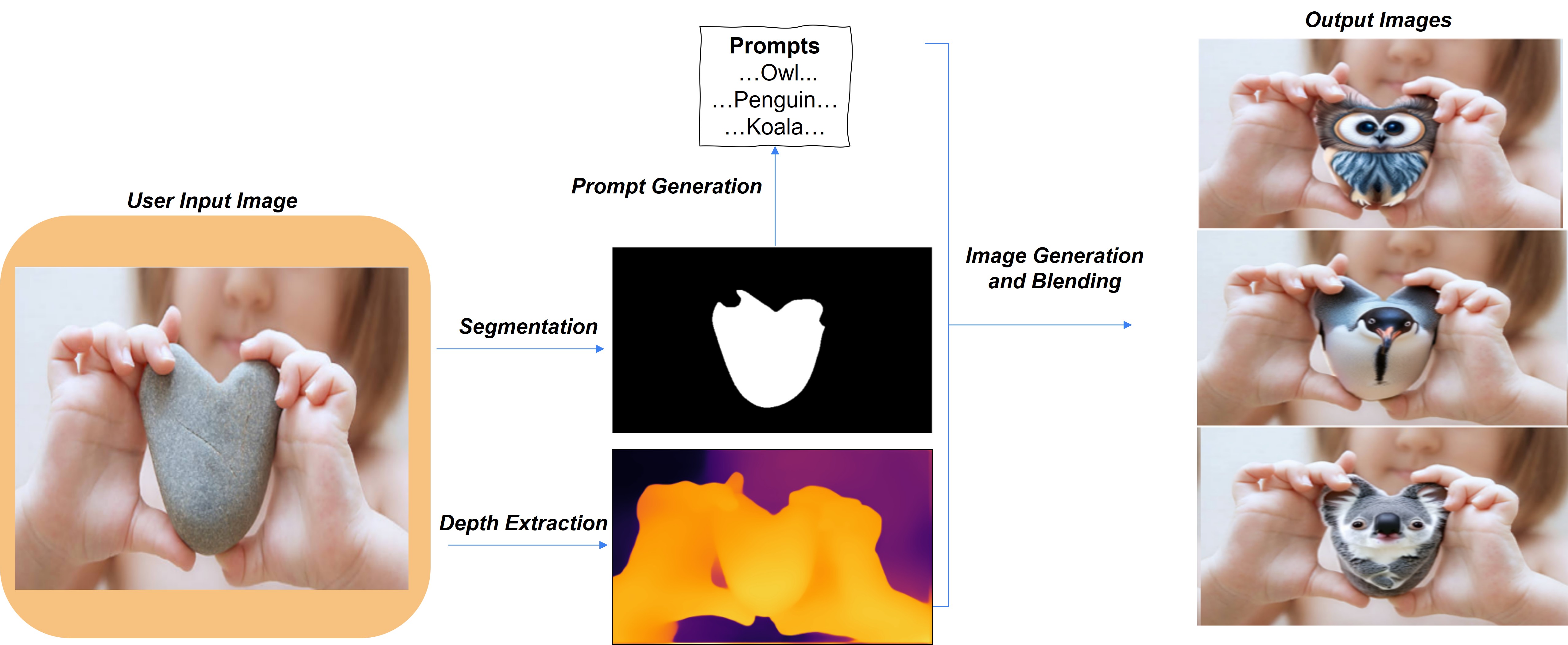}
    \caption{Overview of the Shape2Animal framework. Given a natural object image, the system extracts a salient silhouette, generates candidate animal concepts from its shape, estimates depth for structural guidance, synthesizes a shape-aligned animal image, and blends it into the original scene.}
    \label{fig:overview-pipeline}
\end{figure*}

Figure~\ref{fig:overview-pipeline} illustrate the overview pipeline of Shape2Animal, taking an input natural object image $I_{orig}$ and transforming it into a visually coherent animal form, while preserving the original object's silhouette and background. The transformation process involves four key stages: silhouette segmentation, shape-driven concept interpretation, silhouette-guided animal generation, and image blending.

\begin{figure}[!t]
    \centering
    \includegraphics[width=\linewidth]{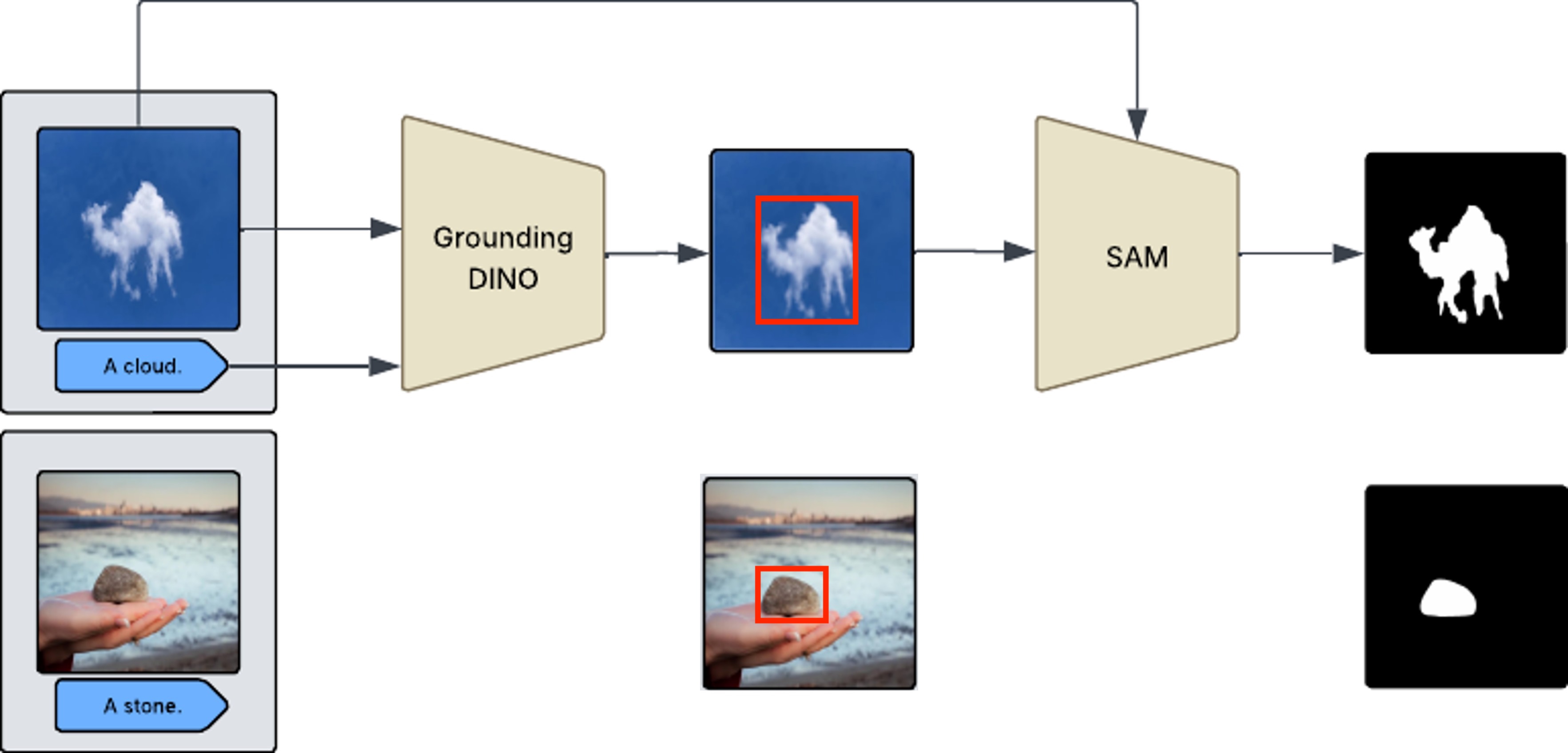}
    \caption{Pipeline of the silhouette segmentation module for extracting the object silhouette mask from the input image.}
    \label{fig:segmentation-fig}
\end{figure}

\subsection{Silhouette Segmentation}

To isolate the target object without requiring manual annotation, we employ a two-step segmentation pipeline (Figure~\ref{fig:segmentation-fig}). First, Grounding DINO~\cite{Zhang:2023:GroundingDINO} detects a set of bounding boxes $\mathcal{B} = \{b_1, b_2, \ldots, b_n\}$ based on open-vocabulary prompts (e.g., ``stone,'' ``cloud,'' ``fire''). Among these, we select the mask $M$ associated with the highest detection confidence. The selected bounding box is then used as input to SAM~\cite{Kirillov:2023:SAM}, which produces a high-quality segmentation mask. 
\begin{equation}
M = \arg\max_{M_i} \, \text{score}_{b_i}.
\end{equation}

This combination of Grounding DINO for language-based localization and SAM for precise mask generation enables robust segmentation of a wide range of natural objects, without requiring domain-specific adaptation.

\subsection{Shape-Driven Concept Interpretation}

\begin{figure}[!t]
    \centering
    \includegraphics[width=\linewidth]{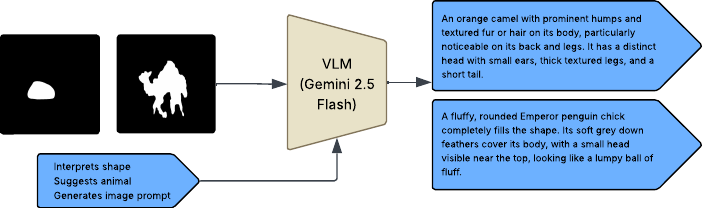}
    \caption{Concept interpreted from silhouette using the Gemini 2.5 model, producing animal names and corresponding descriptive prompts.}
    \label{fig:concept-generation-fig}
\end{figure}

The resulting silhouette mask $M$ is passed to a vision-language model to interpret a plausible animal identity. A structured prompt is formulated to query the Gemini 2.5 API, which interprets the visual shape, classifies the object, and generates a detailed textual description $P$ for image synthesis. An example of this concept interpretation process is shown in Figure~\ref{fig:concept-generation-fig}. The output consists of both a class label (e.g., ``turtle'') and a rendering prompt (e.g., ``A detailed sea turtle filling the shape, patterned green and brown shell texture covers the oval, lower right is a visible grey flipper or tail. No background.''). 

\subsection{Silhouette-Guided Animal Generation}
To generate an animal image that conforms to the silhouette mask, we use a controlled image synthesis pipeline based on the Stable Diffusion XL inpainting model~\cite{podell2023sdxlimprovinglatentdiffusion}. We incorporate \texttt{SDXL-ControlNet-depth} model\footnote{\url{https://huggingface.co/diffusers/controlnet-depth-sdxl-1.0-small}} to guide the image generation using a depth map $D$. The depth map serves as a control input, and we set the strength parameter $\alpha \in \{0.999, 1.0 \} $ to enforce strict adherence to the spatial layout of the original scene. While the text prompt $P$ drives content within the masked region, the ControlNet ensures that the rest of the image remains spatially consistent.
\begin{equation}
    I_{\text{gen}} = \text{ImgGen}(P, M, D),
\end{equation}
where $D=\text{Depth}(I_{\text{orig}})$ is computed from the original image using the depth estimation MiDaS 3.1 model~\cite{birkl2023midas}. Directly applying a high guidance strength in SDXL inpainting may lead to reduced sharpness in the generated image $I_{\text{gen}}$. To address this, we retain the original object region from $I_{\text{orig}}$ and reintroduce it in the blending stage.

\subsection{Image Blending}

In the final stage, we compose the output by combining the inpainted image $I_{\text{gen}}$ with the original input $I_{\text{orig}}$, ensuring seamless integration of the preserved object and the generated background. We blend the masked region using a soft combination of $I_{\text{gen}}$ and $I_{\text{orig}}$ within the foreground mask $M$ at 50\% opacity. The final image $I_{\text{final}}$ is computed as:
\begin{equation}
I_{\text{final}} = 0.5 \cdot (M \odot I_{\text{gen}}) + 0.5 \cdot (M \odot I_{\text{orig}}) + (1 - M) \odot I_{\text{orig}},
\end{equation}
where $\odot$ denotes element-wise multiplication.

\section{Experiments}

\subsection{Dataset}

We collected a total of 62 natural object images spanning three categories: 21 images of stones, 24 images of clouds, and 17 images of fire. Each image was resized to $1024\times1024$ pixels. The dataset was curated to include a diverse range of object shapes and textures within each category.

\mycomment{For the comparative evaluation, we sampled a representative subset of 40 images from this dataset (covering all three categories).}

\subsection{Preliminary Concept Interpretation Study}

Prior to the comparative evaluation, we conducted a focused study to assess the plausibility of the animal concepts generated by the Gemini 2.5 shape-interpretation module in isolation - independent of subsequent image synthesis quality. Nineteen participants (15 male, 4 female; age 20 - 40) evaluated silhouette masks drawn from the full 62-image dataset across two tasks.

In \textit{Task~1 (Concept Matching)}, participants were presented with a segmented object mask only (without any generated image) and asked to freely name the animal they perceived in the shape. The agreement rate between participant responses and the AI-generated labels was 22.63\%, indicating a non-trivial degree of alignment between human intuition and the model's shape-driven concept interpretation.

In \textit{Task~2 (Plausibility Check)}, the same mask was presented together with the AI-predicted animal label, and participants judged whether the suggestion was plausible. 49.67\% of responses rated the AI-predicted animal as plausible, suggesting that nearly half of the generated concepts are convincing to naive viewers even when not independently guessed. Taken together, these results confirm that the shape-interpretation module produces semantically reasonable outputs and that the resulting animal labels constitute a viable basis for downstream image synthesis.

\mycomment{\subsection{Baselines}

We benchmark Shape2Animal against three state-of-the-art general-purpose image generation systems: {ChatGPT} (GPT-4o), {Gemini Pro}, and {Grok}. Since these systems are not architecturally constrained to preserve silhouettes, we applied a standardized zero-drift prompt to all three baselines to enforce the same spatial requirement that Shape2Animal enforces architecturally:
\begin{quote}
\textit{``Perform a `Zero-Drift' image-to-image transformation. Generate a pareidolia effect where the object reveals a hidden animal. Technical Requirement: The Intersection-over-Union (IoU) between the original object's mask and the generated animal must be 1.0. Any generation of pixels outside the original silhouette is a failure. Ensure seamless integration with the existing background.''}
\end{quote}

To ensure a rigorous and consistent comparison, we generated outputs from all four models (Shape2Animal and the three baselines) for each of the 40 images in the evaluation subset, resulting in a total of 160 generated samples for the comparative analysis.}

\begin{figure*}[!t]
    \centering
    \includegraphics[width=\textwidth]{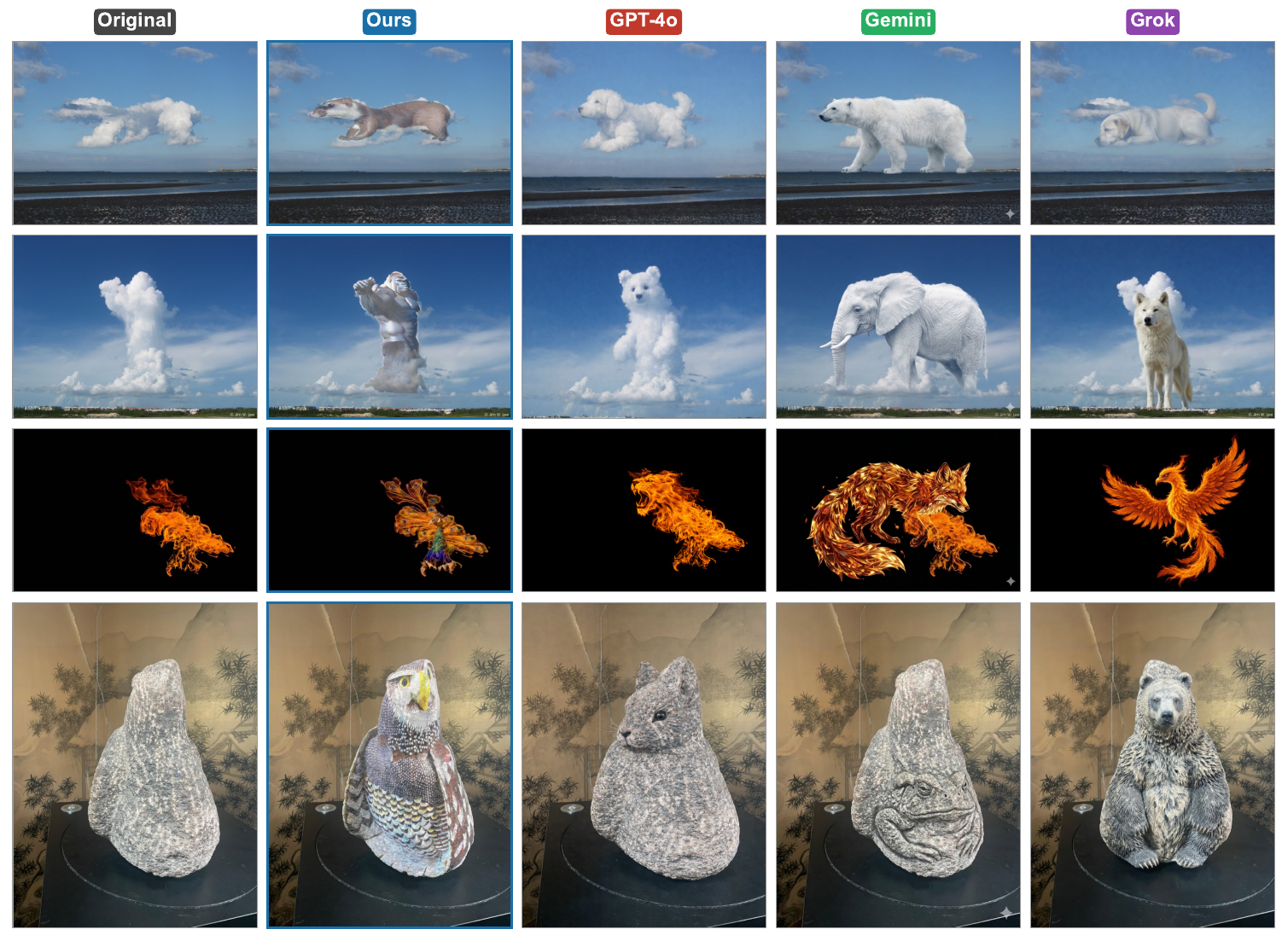}
    \caption{\mycomment{Qualitative comparison of Shape2Animal against baseline models (ChatGPT, Gemini Pro, and Grok) across cloud, stone, and fire categories.}}
    \label{fig:qualitative-comparison}
\end{figure*}

\mycomment{\subsection{Human Perceptual Evaluation}

\subsubsection{Setup}

We recruited 39 participants (21 female, 18 male; age range 16 - 35, majority 21 - 25) with diverse backgrounds including computer science, economics, medicine, communication, psychology, and social sciences. The study was conducted online and anonymized. Method names were hidden and outputs were shuffled per question to prevent order and anchoring effects.

For each input image, participants were shown the four generated outputs in randomized, blinded order and rated each on a 1 - 5 Likert scale across five criteria, guided by a reference rubric provided at the start of the study:
\begin{itemize}
    \item \textbf{Animal Plausibility}: whether the generated image is immediately recognizable as a specific animal.
    \item \textbf{Shape Preservation}: whether the generated animal conforms to the original object's silhouette without deformation.
    \item \textbf{Natural Blending}: whether the animal integrates naturally into the original scene and background.
    \item \textbf{Visual Quality}: perceived sharpness and absence of typical generative artifacts (e.g., distorted anatomy, blurring).
    \item \textbf{Creative Pareidolia}: how cleverly and surprisingly the model exploits the original shape to suggest an animal.
\end{itemize}

Each participant evaluated an average of $18.7 \pm 12.6$ input images, rating all four model outputs per image across all five criteria. Across 39 participants, a total of 729 image-level evaluations were collected, yielding approximately 14,580 individual image-criterion ratings.

\subsubsection{Results}



\begin{figure*}[!t]
    \centering
    \includegraphics[width=\textwidth]{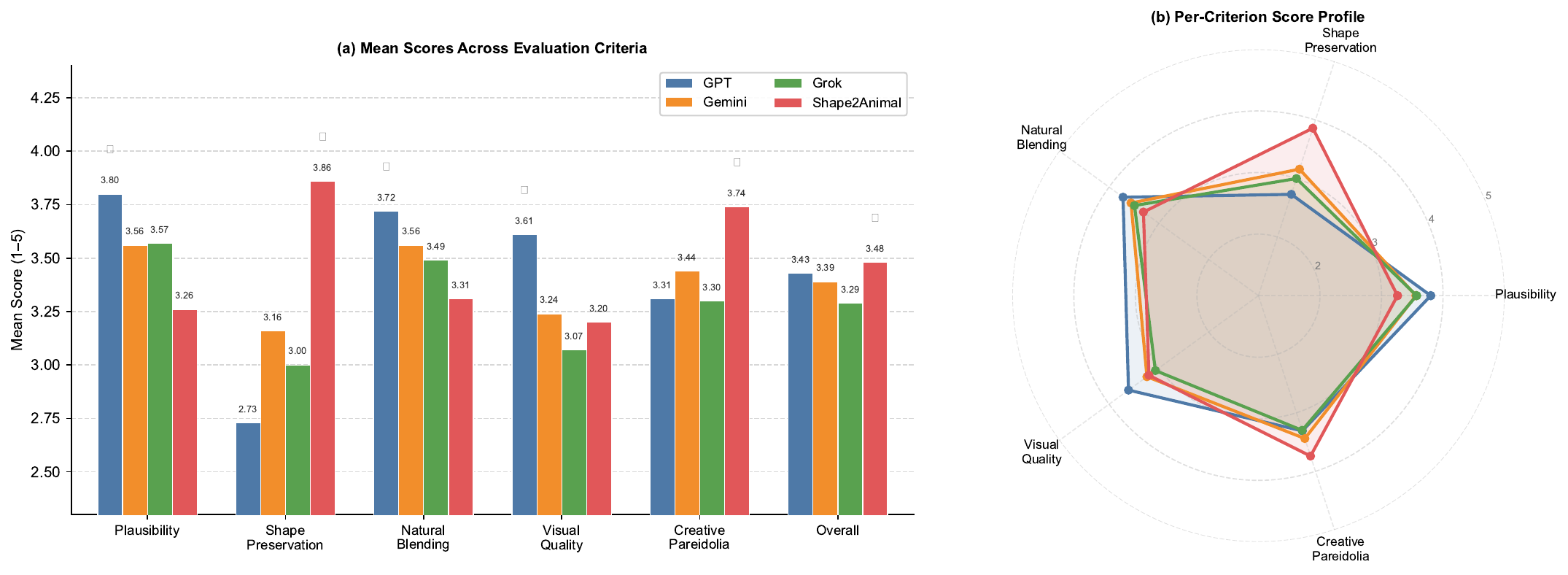}
    \caption{\mycomment{Human perceptual evaluation results. (a) Mean scores across all evaluation criteria for each model. (b) Per-criterion score profile visualized as a radar chart.}}
    \label{fig:userstudy}
\end{figure*}

As shown in Figure~\ref{fig:userstudy}, Shape2Animal achieved the highest Overall Score (3.48) and led on the two criteria most central to the pareidolia task: Shape Preservation (3.86) and Creative Pareidolia (3.74). This outcome reflects a core architectural advantage: Shape2Animal enforces silhouette adherence by design. Accordingly, GPT - despite explicit zero-drift prompting - recorded the lowest Shape Preservation score (2.73) across all models, demonstrating that prompt-level constraints alone are insufficient to guarantee geometric fidelity.

Conversely, GPT ranked highest on Animal Plausibility (3.80), Natural Blending (3.72), and Visual Quality (3.61), reflecting its superior photorealism. This exposes a fundamental trade-off: general-purpose models prioritize visual naturalism over geometric fidelity, producing outputs that appear as convincing animals, yet fail to conform to the original silhouette - the defining requirement of pareidolia generation. 

Set-level win-rate analysis (Table~\ref{tab:winrate}) further corroborates this divergence: Shape2Animal leads on Shape Preservation (66.7\%) and Creative Pareidolia (51.0\%), while GPT dominates on Animal Plausibility (51.0\%), Natural Blending (52.9\%), and Visual Quality (56.9\%).

\begin{table}[t!]
\tablecomment
\caption{Win-rate (\%) per model and criterion - percentage of evaluation 
sets in which each model achieved the highest score. Bold indicates the best 
score per column. Plaus.~=~Animal Plausibility; Shape~=~Shape Preservation; 
Blend~=~Natural Blending; Visual~=~Visual Quality; Creative~=~Creative Pareidolia.}
\label{tab:winrate}
\centering
\setlength{\tabcolsep}{4pt}
\begin{tabular}{lccccc}
\toprule
\textbf{Model} & \textbf{Plaus.} & \textbf{Shape} & \textbf{Blend} & \textbf{Visual} & \textbf{Creative} \\
\midrule
GPT            & \textbf{50.98} & 9.80           & \textbf{52.94} & \textbf{56.86} & 17.65          \\
Gemini         & 27.45          & 15.69          & 23.53          & 23.53          & 17.65          \\
Grok           & 13.73          & 7.84           & 11.76          & 9.80           & 13.73          \\
Shape2Animal   & 7.84           & \textbf{66.67} & 11.76          & 9.80           & \textbf{50.98} \\
\bottomrule
\end{tabular}
\end{table}

A one-way ANOVA confirmed statistically significant differences among the four models across all five criteria  ($p < 0.001$ for all individual metrics). Subgroup analysis further revealed that in-domain evaluators (IT, engineering, design, multimedia) rated Shape Preservation, Visual Quality, and Creative Pareidolia significantly differently from out-of-domain evaluators ($p < 0.05$), while no significant difference was observed on Animal Plausibility.




Figure~\ref{fig:qualitative-comparison} presents representative outputs across all three object categories. Shape2Animal produces animal forms that adhere closely to the source silhouette, whereas baseline models exhibit notable boundary overflow despite explicit zero-drift prompting.}

\mycomment{\subsection{AI-Based Perceptual Evaluation}

To supplement the human perceptual evaluation with a scalable and reproducible reference judgment, we employed \textbf{Claude Sonnet 4.6} as an AI evaluator. This serves as an independent reference that can be applied consistently across all 40 input images for each of the four models (160 outputs in total).

The AI evaluator assessed each generated image on four perceptual criteria using a structured scoring rubric with explicit anchors for each score level:

\begin{itemize}
    \item \textbf{Animal Plausibility}: whether the generated image is immediately recognizable as a specific, identifiable animal.
    \item \textbf{Natural Blending}: how naturally the animal integrates with the background in terms of color, lighting, and scene coherence.
    \item \textbf{Visual Quality}: overall sharpness and realism, penalizing generative artifacts such as anatomical distortion or blurring.
    \item \textbf{Creative Pareidolia}: the degree to which the generation cleverly and convincingly exploits the original shape to evoke an animal.
\end{itemize}

Shape preservation was not included as an AI-evaluated criterion, as vision-language models cannot reliably measure geometric overlap from images alone; this dimension is covered exclusively by the objective IoU analysis (Section \ref{sec:iou})

\begin{table}[t!]
\tablecomment
\caption{AI evaluator mean scores per model (1 - 5 scale, $n=40$). Shape preservation is assessed separately via IoU (Table~\ref{tab:iou}). Bold indicates the best score per column.}
\label{tab:aijudge}
\centering
\begin{tabular}{lccccc}
\toprule
\textbf{Model} & \textbf{Plaus.} & \textbf{Color Blend} & \textbf{Visual} & \textbf{Creative} & \textbf{Overall} \\
\midrule
GPT            & 4.80          & \textbf{3.92} & \textbf{4.12} & \textbf{3.92} & \textbf{4.19} \\
Gemini         & \textbf{4.85} & 3.67          & 4.03          & 3.67          & 4.06 \\
Grok           & 4.88          & 3.62          & 3.98          & 3.48          & 3.99 \\
Shape2Animal   & 4.55          & 3.05          & 3.33          & 3.45          & 3.59 \\
\bottomrule
\end{tabular}
\end{table}

The results in Table~\ref{tab:aijudge} indicate that the AI evaluator consistently favors the baseline models across all criteria, further highlighting Shape2Animal's current limitations in visual fidelity and blending. Notably, the Creative Pareidolia scores, a dimension where Shape2Animal previously outperformed baselines in the human study, are also lower in this evaluation.}

\mycomment{\subsection{Shape Preservation: IoU Analysis}
\label{sec:iou}
Shape preservation is the defining requirement of pareidolia generation. We assess it objectively via Intersection-over-Union (IoU), as perceptual models cannot reliably measure geometric overlap from images alone. For each model output, we re-apply the Grounding DINO + SAM segmentation pipeline using the prompt \textit{``an animal''} to extract the generated animal mask, and compute IoU against the original source mask.

Table~\ref{tab:iou} reports mean IoU and standard deviation for each model across all 40 evaluation images. Shape2Animal achieves substantially higher shape preservation (0.850 $\pm$ 0.265) compared to all baselines (GPT: 0.730 $\pm$ 0.220; Gemini: 0.713 $\pm$ 0.269; Grok: 0.611 $\pm$ 0.277). This gap is consistent with the architectural design of Shape2Animal, which enforces silhouette adherence via a constrained inpainting pipeline, whereas the three baselines must infer the constraint purely from the zero-drift prompt instruction.}

\begin{table}[t!]
\tablecomment
\caption{Shape Preservation measured by IoU (mean $\pm$ std, $n=40$). Higher is better. Shape2Animal enforces shape adherence architecturally; baselines rely on prompt instruction only.}
\label{tab:iou}
\centering
\begin{tabular}{lcc}
\toprule
\textbf{Model} & \textbf{Mean IoU} & \textbf{Std} \\
\midrule
GPT   & 0.730 & 0.220 \\
Gemini & 0.713 & 0.269 \\
Grok  & 0.611 & 0.277 \\
\textbf{Shape2Animal} & \textbf{0.850} & \textbf{0.265} \\
\bottomrule
\end{tabular}
\end{table}

\subsection{Discussion}

\subsubsection{Potential Applications}
\begin{figure}[t!]
    \centering
    \includegraphics[width=\linewidth]{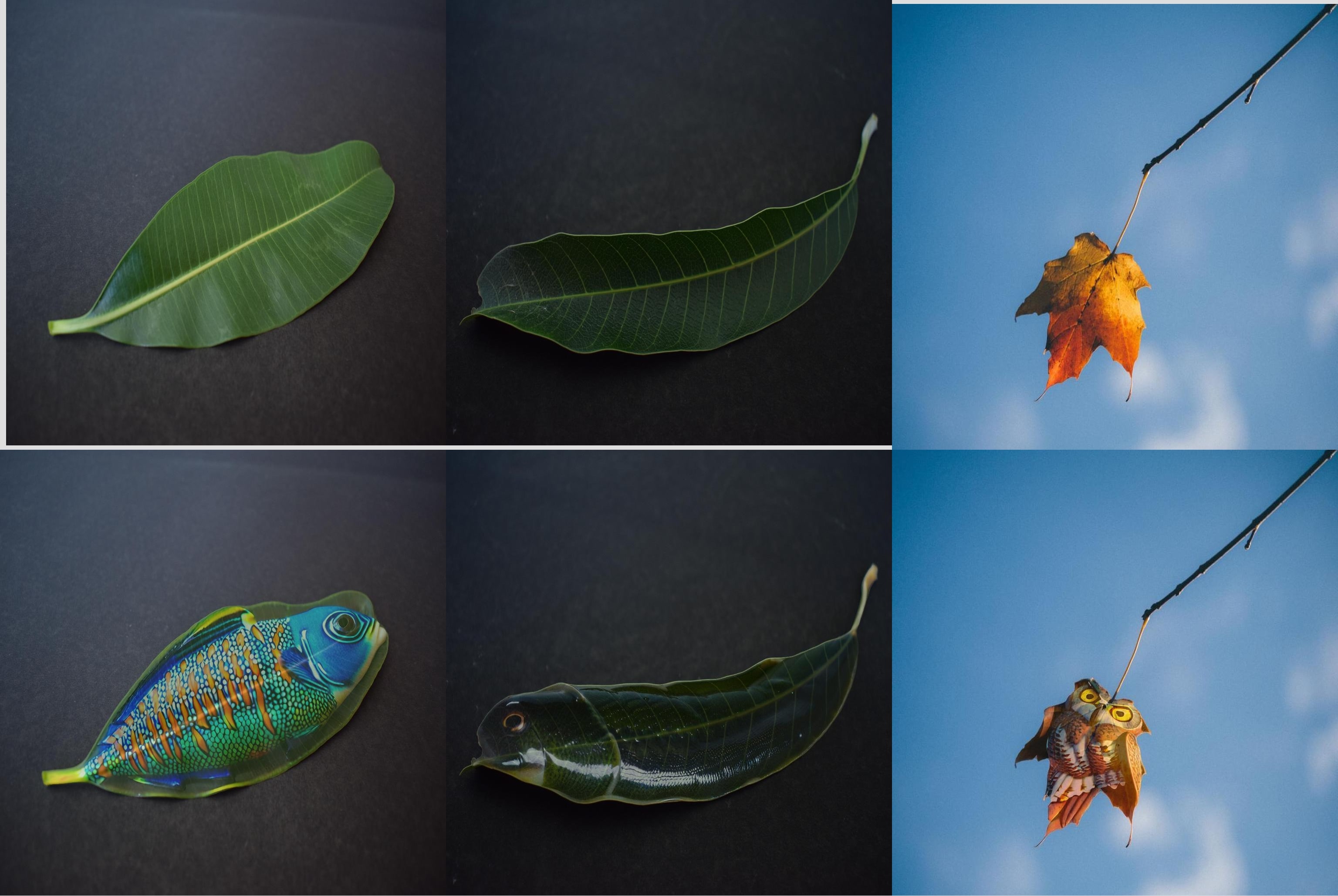}
    \caption{Shape2Animal applied to leaves, demonstrating generalization beyond the three primary categories.}
    \label{fig:leaves-fig}
\end{figure}

The Shape2Animal pipeline also generalizes well to other natural forms, such as foliage. Leaf shapes, with their diverse and organic silhouettes, present promising opportunities for creative interpretation, as shown in Figure~\ref{fig:leaves-fig}. This adaptability opens up broader applications in educational visualization, interactive art, AR content generation, and creative image editing.

\subsubsection{Limitations}

\begin{figure}[t!]
    \centering
    \includegraphics[width=\linewidth]{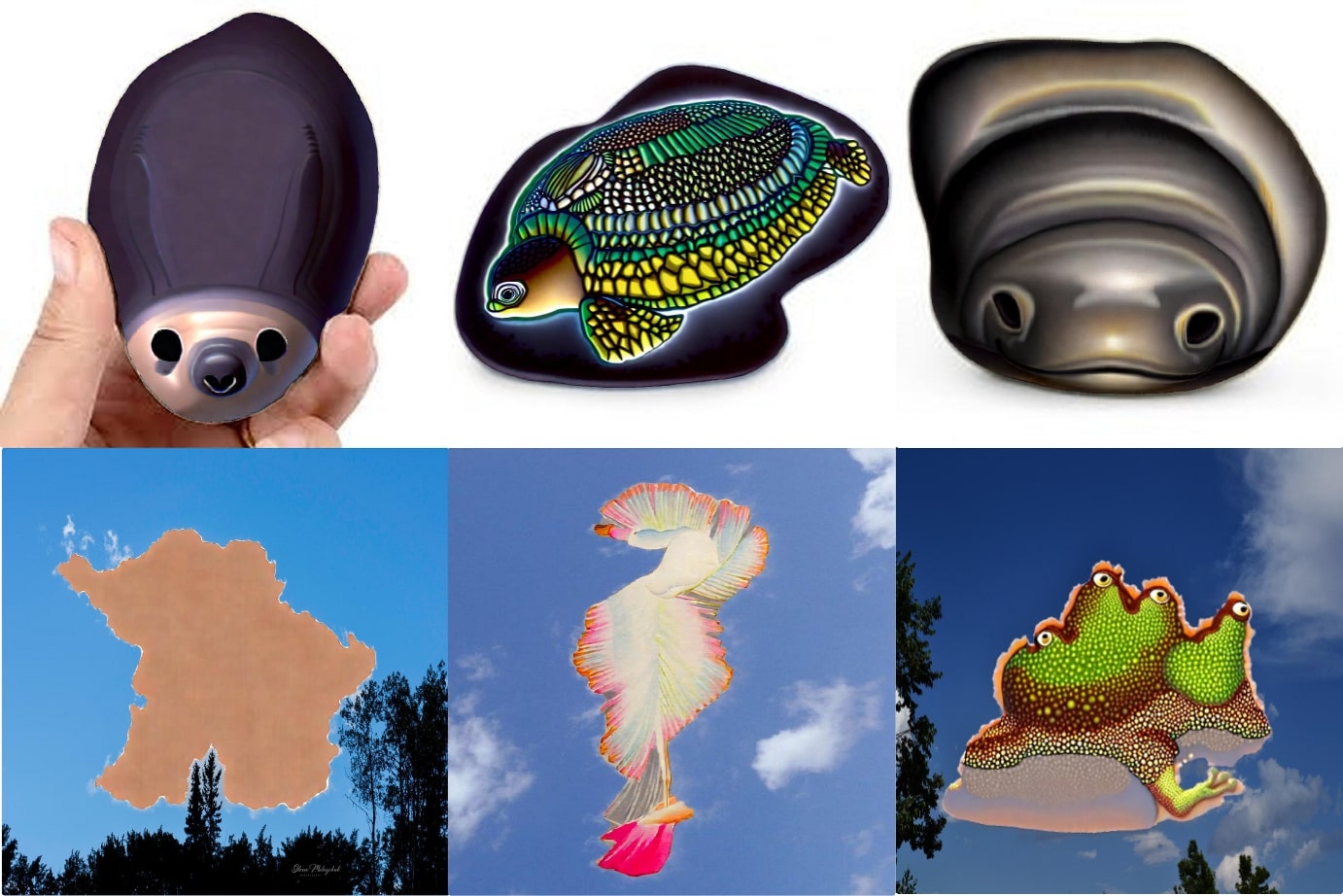}
    \caption{Failure cases of Shape2Animal.}
    \label{fig:diffusion-model-failed}
\end{figure}

Shape2Animal inherits limitations from the foundation models it builds upon. Segmentation failures can occur when SAM~\cite{Kirillov:2023:SAM} struggles with ambiguous or low-contrast objects, leading to incomplete or inaccurate masks. During concept grounding, Gemini may generate implausible animal suggestions when faced with highly abstract shapes. In the synthesis stage, Stable Diffusion XL~\cite{podell2023sdxlimprovinglatentdiffusion} can produce anatomically distorted or incoherent results, especially when the silhouette diverges significantly from familiar forms. These failure cases (Figure~\ref{fig:diffusion-model-failed}), ranging from incorrect segmentation and implausible concept mapping to flawed image generation, highlight key challenges in end-to-end visual reasoning and point to areas for future improvement.

\mycomment{The comparative evaluation further revealed a core trade-off inherent to Shape2Animal's architecture: strict silhouette enforcement, while ensuring the highest shape preservation scores (IoU: 0.850; human Shape Preservation: 3.86), constrains the synthesis model and reduces Animal Plausibility (3.26 vs.\ GPT's 3.80) and the AI-judged natural blending quality. Statistically significant differences in perception were observed between in-domain and out-of-domain evaluators on Shape Preservation, Visual Quality, and Creative Pareidolia ($p < 0.05$), suggesting that domain expertise affects how certain visual and structural qualities are judged. The participant pool, while more demographically balanced than prior work (39 participants, 54\% female), remains skewed toward young adults (79.5\% aged 21 - 25), which may limit generalizability to broader populations. 

Furthermore, the divergence between human and AI evaluations is noteworthy: while human participants rated Shape2Animal ighest on Creative Pareidolia, the AI evaluator consistently favored baseline models across all criteria, suggesting that vision-language models used as judges may share a photorealism bias with general-purpose generative models, potentially undervaluing shape-adherence qualities central to the pareidolia task.}

\section{Conclusion}

\mycomment{This paper introduced Shape2Animal, a novel framework for transforming natural object silhouettes into animal figures through a multi-stage pipeline that combines precise segmentation, vision-language-based prompt generation, shape-constrained image synthesis, and seamless background integration. A comprehensive evaluation including objective IoU measurement, a human perceptual evaluation (N=39), and an AI-based evaluation demonstrates that Shape2Animal uniquely excels at the shape-preservation requirement central to pareidolia generation, while general-purpose models prioritize visual realism at the cost of geometric fidelity. By offering an intuitive yet powerful tool for artists, educators, and designers, this work expands the possibilities of visual storytelling and generative media. Future work may involve integrating more advanced vision-language models, improving the blending stage for stronger material integration, and fine-tuning components for better domain-specific performance.}

\section*{Acknowledgment}

This research is funded by Vietnam National Foundation for Science and Technology Development (NAFOSTED) under Grant Number 102.05-2023.31.

\bibliographystyle{IEEEtran}
\bibliography{IEEEabrv,ref}

\begin{thebibliography}{10}
\providecommand{\url}[1]{#1}
\csname url@samestyle\endcsname
\providecommand{\newblock}{\relax}
\providecommand{\bibinfo}[2]{#2}
\providecommand{\BIBentrySTDinterwordspacing}{\spaceskip=0pt\relax}
\providecommand{\BIBentryALTinterwordstretchfactor}{4}
\providecommand{\BIBentryALTinterwordspacing}{\spaceskip=\fontdimen2\font plus
\BIBentryALTinterwordstretchfactor\fontdimen3\font minus \fontdimen4\font\relax}
\providecommand{\BIBforeignlanguage}[2]{{%
\expandafter\ifx\csname l@#1\endcsname\relax
\typeout{** WARNING: IEEEtran.bst: No hyphenation pattern has been}%
\typeout{** loaded for the language `#1'. Using the pattern for}%
\typeout{** the default language instead.}%
\else
\language=\csname l@#1\endcsname
\fi
#2}}
\providecommand{\BIBdecl}{\relax}
\BIBdecl

\bibitem{Bednarik2024CollectiveP}
R.~Bednarik, ``Collective pareidolia,'' \emph{Qeios}, 2024.

\bibitem{Liu2014FacePareidolia}
J.~Liu, J.~Li, L.~Feng, L.~Li, J.~Tian, and K.~Lee, ``Seeing jesus in toast: Neural and behavioral correlates of face pareidolia,'' \emph{Cortex}, vol.~53, pp. 60--77, 2014.

\bibitem{Doniger2012AnimalsImagination}
W.~Doniger, \emph{Animals and the Human Imagination: A Companion to Animal Studies}.\hskip 1em plus 0.5em minus 0.4em\relax Columbia University Press, 2012.

\bibitem{MartinezConde2015BrainSeesFaces}
S.~Martinez-Conde, ``The brain sees faces everywhere,'' \emph{Scientific American}, Sep 2015.

\bibitem{Ramachandran1999ArtBrain}
V.~S. Ramachandran and W.~Hirstein, ``The science of art: A neurological theory of aesthetic experience,'' \emph{Journal of Consciousness Studies}, vol.~6, no. 6--7, pp. 15--51, 1999.

\bibitem{Somaini2022OnTA}
A.~Somaini, ``On the altered states of machine vision,'' \emph{AN-ICON. Studies in Environmental Images [ISSN 2785-7433]}, 2022.

\bibitem{podell2023sdxlimprovinglatentdiffusion}
D.~Podell, Z.~English, K.~Lacey, A.~Blattmann, T.~Dockhorn, J.~M{\"u}ller, J.~Penna, and R.~Rombach, ``Sdxl: Improving latent diffusion models for high-resolution image synthesis,'' \emph{arXiv preprint arXiv:2307.01952}, 2023.

\bibitem{saharia2022photorealistictexttoimagediffusionmodels}
C.~Saharia, W.~Chan, S.~Saxena, L.~Li, J.~Whang, E.~L. Denton, K.~Ghasemipour, R.~Gontijo~Lopes, B.~Karagol~Ayan, T.~Salimans \emph{et~al.}, ``Photorealistic text-to-image diffusion models with deep language understanding,'' \emph{Advances in neural information processing systems}, vol.~35, pp. 36\,479--36\,494, 2022.

\bibitem{zhu2020unpairedimagetoimagetranslationusing}
J.-Y. Zhu, T.~Park, P.~Isola, and A.~A. Efros, ``Unpaired image-to-image translation using cycle-consistent adversarial networks,'' in \emph{IEEE international conference on computer vision}, 2017, pp. 2223--2232.

\bibitem{zhang2023addingconditionalcontroltexttoimage}
L.~Zhang, A.~Rao, and M.~Agrawala, ``Adding conditional control to text-to-image diffusion models,'' in \emph{IEEE/CVF international conference on computer vision}, 2023, pp. 3836--3847.

\bibitem{Zhang:2023:GroundingDINO}
S.~Zhang, X.~Chen, X.~Wang, Z.~Liu, S.~Liu, M.~Li, and P.~Luo, ``Grounding dino: Marrying dino with grounded pre-training for open-set object detection,'' \emph{arXiv preprint arXiv:2303.05499}, 2023, accessed April 2025.

\bibitem{Kirillov:2023:SAM}
A.~Kirillov, E.~Mintun, N.~Ravi, H.~Mao, C.~Rolland, L.~Gustafson, T.~Xiao, S.~Whitehead, A.~C. Berg, W.-Y. Lo, P.~Doll{\'a}r, and R.~Girshick, ``Segment anything,'' \emph{arXiv preprint arXiv:2304.02643}, 2023, accessed April 2025.

\bibitem{goodfellow2014generativeadversarialnetworks}
I.~Goodfellow, J.~Pouget-Abadie, M.~Mirza, B.~Xu, D.~Warde-Farley, S.~Ozair, A.~Courville, and Y.~Bengio, ``Generative adversarial networks,'' \emph{Communications of the ACM}, vol.~63, no.~11, pp. 139--144, 2020.

\bibitem{Rombach_2022_CVPR}
R.~Rombach, A.~Blattmann, D.~Lorenz, P.~Esser, and B.~Ommer, ``High-resolution image synthesis with latent diffusion models,'' in \emph{IEEE/CVF Conference on Computer Vision and Pattern Recognition (CVPR)}, June 2022, pp. 10\,684--10\,695.

\bibitem{ramesh2022hierarchicaltexttoimage}
A.~Ramesh, P.~Dhariwal, A.~Nichol, C.~Chu, and M.~Chen, ``Hierarchical text-conditional image generation with clip latents,'' \emph{arXiv preprint arXiv:2204.06125}, vol.~1, no.~2, p.~3, 2022.

\bibitem{mou2023t2iadapter}
L.~Mou \emph{et~al.}, ``T2i-adapter: Learning adapters to dig out more controllable ability for text-to-image diffusion models,'' in \emph{IEEE/CVF Conference on Computer Vision and Pattern Recognition (CVPR)}, 2023, pp. 12\,345--12\,354.

\bibitem{radford2021learningtransferable}
A.~Radford, J.~W. Kim, C.~Hallacy, A.~Ramesh, G.~Goh, S.~Agarwal, G.~Sastry, A.~Askell, P.~Mishkin, J.~Clark, G.~Krueger, and I.~Sutskever, ``Learning transferable visual models from natural language supervision,'' in \emph{38th International Conference on Machine Learning}, vol. 139.\hskip 1em plus 0.5em minus 0.4em\relax PMLR, 2021, pp. 8748--8763.

\bibitem{alayrac2022flamingofewshot}
J.-B. Alayrac, J.~Donahue, P.~Luc, A.~Miech, I.~Barr, Y.~Hasson, K.~Lenc, A.~Mensch, K.~Millican, M.~Reynolds \emph{et~al.}, ``Flamingo: a visual language model for few-shot learning,'' \emph{Advances in neural information processing systems}, vol.~35, pp. 23\,716--23\,736, 2022.

\bibitem{birkl2023midas}
R.~Birkl, D.~Wofk, and M.~M{\"u}ller, ``Midas v3.1 -- a model zoo for robust monocular relative depth estimation,'' \emph{arXiv preprint arXiv:2307.14460}, 2023.

\end{thebibliography}

\begin{IEEEbiographynophoto}{Quoc-Duy Tran} is an undergraduate student at the University of Science, Ho Chi Minh City, Vietnam. His research interests include machine learning, computer vision, and mixed reality.
\end{IEEEbiographynophoto}

\begin{IEEEbiographynophoto}{Anh-Tuan Vo} is an undergraduate student at the University of Science, Ho Chi Minh City, Vietnam. His research interests include machine learning, multimedia, and human-computer interaction.
\end{IEEEbiographynophoto}

\begin{IEEEbiographynophoto}{Dinh-Khoi Vo} is currently a graduate student at the University of Science, Ho Chi Minh City, Vietnam. He received the B.Sc. degree in software engineering from the University of Science in 2024. His research interests include machine learning, computer vision, and human-computer interaction.
\end{IEEEbiographynophoto}

\begin{IEEEbiographynophoto}{Tam V. Nguyen} is currently a professor at the University of Dayton, Ohio, United States. He received the Ph.D. degree in computer science from the National University of Singapore in 2013. His research interests include computer vision, deep learning, and mixed reality.
\end{IEEEbiographynophoto}

\begin{IEEEbiographynophoto}{Minh-Triet Tran} is currently a professor at the University of Science, Ho Chi Minh City, Vietnam. He received the Ph.D. degree in computer science from the University of Science in 2009. His research interests include machine learning, computer vision, and multimedia.
\end{IEEEbiographynophoto}

\begin{IEEEbiographynophoto}{Trung-Nghia Le} is currently a senior researcher and lecturer at the University of Science, Ho Chi Minh City, Vietnam. He received the Ph.D. degree in computer science from the National Institute of Informatics, Japan, in 2018. His research interests include machine learning, computer vision, and multimedia.
\end{IEEEbiographynophoto}

\end{document}